%% file: main.tex
\documentclass[sigconf,fleqn]{article}

\usepackage{booktabs} 
\usepackage{subcaption}
\usepackage{multirow}
\usepackage{graphicx}

\newcommand{\eg}{{\it e.g.}}
\newcommand{\ie}{{\it i.e.}}
\newcommand{\etal}{{\it et al. }}

\begin{document}
\title{Large Scale Evolution of Convolutional Neural Networks Using Volunteer Computing}

\author{Travis Desell\thanks{University of North Dakota. 3950 Campus Road, Grand Forks, North Dakota, 58201.}}

\maketitle

\begin{abstract}
    \input{00-abstract}
\end{abstract}

\input{01-introduction.tex}

\input{02-exact.tex}
\input{03-implementation.tex}
\input{04-results.tex}
\input{05-discussion.tex}
\input{06-acknowledgements.tex}

\bibliographystyle{abbrv}
\bibliography{../references} 

\end{document}

%% file: 00-abstract.tex
This work presents a new algorithm called evolutionary exploration of augmenting convolutional topologies (EXACT), which is capable of evolving the structure of convolutional neural networks (CNNs).  EXACT is in part modeled after the neuroevolution of augmenting topologies (NEAT) algorithm, with notable exceptions to allow it to scale to large scale distributed computing environments and evolve networks with convolutional filters. In addition to multithreaded and MPI versions, EXACT has been implemented as part of a BOINC volunteer computing project, allowing large scale evolution.  During a period of two months, over 4,500 volunteered computers on the Citizen Science Grid trained over 120,000 CNNs and evolved networks reaching 98.32\% test data accuracy on the MNIST handwritten digits dataset.  These results are even stronger as the backpropagation strategy used to train the CNNs was fairly rudimentary (ReLU units, L2 regularization and Nesterov momentum) and these were initial test runs done without refinement of the backpropagation hyperparameters. Further, the EXACT evolutionary strategy is independent of the method used to train the CNNs, so they could be further improved by advanced techniques like elastic distortions, pretraining and dropout.  The evolved networks are also quite interesting, showing "organic" structures and significant differences from standard human designed architectures.

%% file: 01-introduction.tex
\section{Introduction}
\label{sec:introduction}

Convolutional Neural Networks (CNNs) have become a highly active area of research due to strong results in areas such as image classification~\cite{krizhevsky2012imagenet,lecun1998gradient}, video classification~\cite{karpathy2014large}, sentence classification~\cite{kim2014convolutional}, and speech recognition~\cite{hinton2012deep}, among others. Significant progress has been made in the design of CNNs, from the venerable LeNet 5~\cite{lecun1998gradient} to more recent large and deep networks such as AlexNet~\cite{krizhevsky2012imagenet}, VGGNet~\cite{simonyan2014very}, GoogleNet~\cite{szegedy2015going} and ResNet~\cite{he2016deep}, however less work has been made in the area of automated design of CNNs.

There exist a number of neuroevolution techniques capable of evolving the structure of feed forward and recurrent neural networks, such as NEAT~\cite{stanley2002evolving}, HyperNEAT~\cite{stanley2009hypercube}, CoSyNE~\cite{gomez2008accelerated}, as well as ant colony optimization based approaches~\cite{desell-evostar-2015,salama2014novel}. However, these have not yet been applied to CNNs due to the size and structure of CNNs, not to mention the significant amount of time required to train one.

Koutn\'{i}k \etal have published a work titled {\it Evolving Deep Unsupervised Convolutional Networks for Vision-Based Reinforcement Learning}~\cite{koutnik2014evolving}, however in this work the structure of the CNN used was held fixed while only a small recurrent neural network controller (which takes output from the CNN) was evolved using the CoSyNE~\cite{gomez2008accelerated} algorithm.  To the authors' knowledge, the only work that comes close to evolving the structure of CNNs is a yet to be published work by Zoph \etal~\cite{zoph2016neural}, which uses a recurrent neural network trained with reinforcement learning to maximize the expected accuracy of generated architectures on a validation set of images.  However, this approach is gradient based and generates CNNs by layer, with each layer having a fixed filter width and height, stride width and height, and number of filters.

Highlighting the cutting edge nature of this topic, between the time of this works submission and notification, recent preprints have been submitted to {\tt arXiv.org}. Xie \etal propose a Genetic CNN method which encodes CNNs as binary strings~\cite{xie2017geneticcnn}, however they only evolve structure of convolution operations between pooling layers, and keep the filter sizes fixed. Miikkulainen \etal have proposed CoDeepNEAT, which is also based on NEAT with each node acts as an entire layer, and the type of layer and hyperparameters are evolved~\cite{mikkulainen2017codeepneat}. However, connections within layers are fixed depending on their type without arbitrary connections.  Real \etal, have also evolved image classifiers on the CIFAR-10 and CIFAR-100 datasets in work most close to this~\cite{real2017evolution}. They also use a distributed algorithm to evolve progressively more complex CNNs through mutation operations, and handle conflicts in filter sizes by reshaping non-primary edges with zeroth order interpolation.

This work presents a new algorithm, Evolutionary Exploration of Augmenting Convolutional Topologies (EXACT), which can evolve CNNs of arbtrary structure and filter size. Due to high computational demands, it has been implemented as part of the Citizen Science Grid\footnote{http://csgrid.org}, a Berkeley Open Infrastructure for Network Computing (BOINC)~\cite{anderson_boinc_2005} volunteer computing project. Using the MNIST handwritten digits dataset~\cite{lecun1998mnist} as a benchmark, over 4,500 volunteered computers trained over 120,000 CNNs and achieved networks with 98.32\% accuracy on the test data. These results are made stronger due to the fact that the CNNs were trained with a relatively rudimentary strategy (ReLU units, L2 regularization and Nesterov momentum) and the fact that these were preliminary test runs done without fine tuning hyperparameters. When compared to a set of human designed CNNs without max pooling and trained with the same backpropagation implementation and hyperparameters, the evolved CNNs showed improved training and test data accuracy, and significantly reduced training and testing error. Further, the evolved CNNs show interesting structures and show significant differences from human designed CNNs seen in the literature, which the author hopes may inform or spark new insight in the design of future CNN architectures.





%% file: 02-exact.tex
\section{Evolutionary Exploration of Augmenting Convolutional Topologies}
\label{sec:exact}

The EXACT algorithm starts with the observation that any two filters of any size within a CNN can be connected by a convolution of size $conv_d = |out_d - in_d| + 1$, where $out_d$ and $in_d$ are the size of the output and input filters, respectively, and $conv_d$ is the size of the convolution in dimension $d$. The consequence of this observation is that the structure of a CNN can be evolved solely by determining the sizes of the filters and how they are connected. Instead of evolving the weights of individual neurons and how they are connected, as done in the NEAT~\cite{stanley2002evolving} algorithm, the architecture of a CNN can be evolved in a similar fashion except on the level of how filters are connected, with additional operators to modify the filter sizes.  

Due to the computational expense of training CNNs, EXACT has been designed with scalable distributed execution in mind. It uses an asynchronous evolution strategy, which has been shown by Desell \etal to allow scalability up to potentially millions of compute hosts in a manner independent of population size~\cite{desell-analysis-massive-eas-2010}. A master process manages a population of \emph{CNN genomes} (the filter sizes and how they are connected) along with their fitness (the minimized error after backpropagation on that CNN). Worker processes request CNN genomes to evaluate from the master, which generates them either through applying mutation operations to a randomly selected genome in the population (see Section~\ref{sec:mutation}) or by selecting two parents and performing crossover to generate a child genome (see Section~\ref{sec:crossover}).  When that worker completes training the CNN, it reports the CNN along with its fitness back to the master, which will insert it into the population and remove the least fit genome if it would improve the population.  This asynchronous approach has an additional benefit in that the evolved CNNs have different training times, and no worker need wait in the results of another to request another CNN to evaluate, \ie, this approach automatically load balances itself.

Given the strong advances made by in the machine learning community for training deep CNNs and the sheer number of weights in large CNNs, attempts to try and evolve the weights in the neural networks did not seem feasible.  Instead, EXACT allows for any CNN training method to be plugged in to perform the fitness evaluation done by the workers.  In this way, EXACT can benefit from further advances by the machine learning community and also make for an interesting platform to evaluate different neural network training algorithms.

\subsection{Population Initialization}
\label{sec:population_initialization}

Generation of the initial population starts with first generating a \emph{minimal CNN genome}, which consists solely of the input node, which is the size of the training images (plus padding if desired), and one output node per training class for a softmax output layer.  In this case of this work which uses the MNIST handwritten digits dataset, this is a 28x28 input node, and 10 output nodes -- this also happens to be the simplest benchmark NNs used to evaluate results (see Figure~\ref{fig:one_layer_nn}).  This is sent as the CNN genome for the first work request, and also inserted into the population with $\infty$ as fitness, denoting that it had not been evaluated yet. Further work requests are fulfilled by taking a random member of the population (which will be initially just the minimal CNN genome), mutating it, inserting the mutation into the population with $\infty$ as fitness and sending that CNN genome to the worker to evaluate.  Once the population has reached a user specified population size through inserting newly generated mutations and results received by workers, work requests are fulfilled by either mutation or crossover, depending on a user specified crossover rate (\eg, a 20\% crossover rate will result in 80\% mutation).

\subsection{Mutation Operations}
\label{sec:mutation}

When a CNN genome is selected for mutation, a user specified number of the following mutations are performed. In testing, it was found to be beneficial to allow for greater variation in the CNNs generated. Each operator is selected with a user specified rate.  Currently, the CNNs evolved do not utilize pooling layers, however modifying the size of max pooling done by the filters is an area of future work to be implemented as an additional mutation operation.

The operations performed are similar to the NEAT algorithm, with the addition of operations to change the node size. Additionally, whereas NEAT only requires innovation numbers for new edges, EXACT requires innovation numbers for both new nodes and new edges.  The master process keeps track of all node and edge innovations made, which is required to perform the crossover operation in linear time without a graph matching algorithm.

\paragraph{Disable Edge} This operation randomly selects an enabled edge in a CNN genome and disables it so that it is not used.  The edge remains in the genome. As the \emph{disable edge} operation can potentailly make an output node unreachable, after all mutation operations have been performed to generate a child CNN genome, if any output node is unreachable that CNN genome is discarded and a new child is generation by another attempt at mutation.

\paragraph{Enable Edge} If there are any disabled edges in the CNN genome, this operation selects a disabled edge at random and enables it.

\paragraph{Split Edge} This operation selects an enabled edge at random and disables it. It creates a new node (creating a new node innovation) and two new edges (creating two new edge innovations), and connects the input node of the split edge to the new node, and the new node to the output node of the split edge. The filter size of the new node is set to $\frac{i_x + o_x}{2.0}$ by $\frac{i_y + o_y}{2.0}$, where $i_d$ and $o_d$ are the size of the input and output filters, respectively, in dimension $d$ (\ie, the size of the new node is halfway between the size of the input and output nodes).  Further, the new node is given a depth value, $depth_{new} = \frac{depth_{output} + depth_{input}}{2.0}$, which is used by the \emph{add edge} operation and to linearly perform forward and backward propagation without graph traversal.

\paragraph{Add Edge} This operation selects two nodes $n_1$ and $n_2$ within the CNN Genome at random, such that $depth_{n_1} < depth_{n_2}$ and such that there is not already an edge between those nodes in this CNN Genome, and then adds an edge from $n_1$ to $n_2$. This ensures that all edges generated feed forward (currently EXACT does not evolve recurrent CNNs). If an edge between $n_1$ and $n_2$ exists within the master's innovation list, that edge innovation is used, otherwise this creates a new edge innovation.

\paragraph{Change Node Size} This operation selects a node at random from within the CNN Genome and randomly increases or decreases its filter size in both the x and y dimension. For this work, the potential size modifications used were [-2, -1, +1, +2].

\paragraph{Change Node Size X} This operation is the same as \emph{change node size} except that it only changes the filter size in the x dimension.

\paragraph{Change Node Size Y} This operation is the same as \emph{change node size} except that it only changes the filter size in the y dimension.

\subsection{Crossover}
\label{sec:crossover}

Crossover utilizes two hyperparameters, the \emph{more fit parent crossover rate} and the \emph{less fit parent crossover rate}. Two parent CNN genomes are selected, and the child CNN genome is generated from every edge that appears in both parents. Edges that only appear in the more fit parent are added randomly at the \emph{more fit parent crossover rate}, and edges that only appear in the less fit parent are added randomly at the \emph{less fit parent crossover rate}. Edges not added by either parent are also carried over into the child CNN genome, however they are set to disabled. Nodes are then added for each input and output of an edge.  If the more fit parent has a node with the same innovation number, it is added from the more fit parent (\ie, filter sizes are inherited from the more fit parent if possible), and from the less fit parent otherwise.

\subsection{Epigenetic Weight Initialization}
\label{sec:epigenetic_weights}

While EXACT is independent of the method used to train CNNs, it does however present an interesting opportunity for weight initialization. As after the initial population is evaluated, child genomes are generated from one or two trained parent CNNs. \emph{Can the weights from unmutated nodes and edges be reused as a better starting point for backpropagation?} The EXACT implementation optionally allows for weights of the parent CNN genomes to be carried over into child genomes, as \emph{"epigenetic" weight initialization} -- as these weights are a modification of how the genome is expressed as opposed to a modification of the genome itself.

%% file: 03-implementation.tex
\section{EXACT on Volunteered Hosts}
\label{sec:implementation}

The EXACT source code has been made freely available as an open source project on GitHub\footnote{https://github.com/travisdesell/exact}.  It has a multithreaded implementation for small scale use, and an MPI implementation for use on high performance computing clusters. However, to provide enough computational resources to perform EXACT on a large scale, it was also implemented as part of a BOINC project. BOINC clients running on volunteered hosts serve as worker processes, and server side daemons were developed to validate results and handle the master EXACT process. This required developing code to save the state of the EXACT master process to a MySQL database such that these server side daemons could be stopped and restarted without loss of progress, which enabled checkpointing in the multithreaded and MPI versions as well. Due to space limitations, the author recommends~\cite{anderson_boinc_2005,anderson-volunteer-computing-2010} for further details on the BOINC architecture.

In able to be able to utilize BOINC as a platform for training evolved CNNs, a number of implementation challenges needed to be addressed. First, BOINC requires applications to be written in C++ so they can link against the required BOINC libraries to be executed within by the BOINC clients running on volunteered hosts. While there is some work in enabling the use of Python applications in BOINC~\cite{heien2009pymw} or applications within virtual machines~\cite{rentala-vmwrapper}, these efforts are still under development. The requirement of C++, along with other techical requirements described later in this section, precluded the use of popular CNN training packages such as Caffe~\cite{jia2014caffe} or Theano~\cite{bergstra2011theano,bastien2012theano}.

Perhaps the most significant techincal challenge was that in order to prevent users from reporting incorrect or malicious results, each \emph{workunit} needed to be sent to multiple hosts so that the results can be validated against each other. Results which do not match those from other hosts are flagged as invalid and discarded. Only results that are valid award that particular user credit, which can be used to generate currency for Gridcoin whitelisted projects~\cite{gridcoin-2017}, or as motivation to climb various leaderboards~\cite{boincstats-2017}. The applications that run on the hosts must also be able to checkpoint and resume from previously saved states, as users expect to not lose progress if they need to temporarily turn off their BOINC client, or restart or shut down their computers.

For many applications, checkpointing and validation can be easily addressed, however in the case of training CNNs, even slightly different values can accumulate leading to dramatically different results. This means two hosts could be doing valid work, but return results that are incomparable. This also necessitates care being taken by checkpoints.  If the CNN weights and other training values aren't saved and restored with the exact same precision, these small differences can result in significantly different final results.

\subsection{Validation}
\label{sec:validation}

The EXACT code was able to accomplish identical results on a wide variety of BOINC computing hosts. In the current EXACT codebase, Windows hosts from XP to Windows 10 on \emph{Intel86} and \emph{x86\_64} architectures, along with Mac OS X and Linux hosts on \emph{x86\_64} architectures are all returning practically identical results, allowing validation. As the Windows applications were compiled with Visual Studio 2015, the Linux applications were compiled with the Gnu C++ Compiler (g++) and the Mac OS X applications were compiled with the LLVM compiler, inconsistent implementations of the C and C++ Standard Libraries needed to be addressed.

First, at the end of each epoch, the order of training images needs to be shuffled. While the \emph{minstd\_rand0} random number generator returned indentical results across platforms, the C++ \emph{shuffle} function did not due to differing implementations of the \emph{uniform\_distribution} classes. Due to this, EXACT utilizes a custom built Fisher Yates Shuffle function which produces identical results across hosts. Second, on the different platforms, the \emph{exp()} function began to return slightly different results for more extreme input values, resulting in widely varying final results. EXACT uses a custom built \emph{exp()} based on a Taylor Series expansion, which while slower than the C Standard Library implementations, returns identical results across the different platforms.  As the \emph{exp()} function was only used 10 times per image as part of the softmax layer, this did not have any noticable effect on performance.

\subsection{Checkpointing}
\label{sec:checkpointing}

Due to issues involved in robustly developing portable binary files, checkpointing of the CNNs in EXACT is done using a text file. However, similarly to the \emph{exp()} function, printing double precision variables to the checkpoint file with arbitrary precision lead to some data loss and divergent results. This issue was solved by the use of \emph{hexfloats}, which allow full precision I/O of floating point variables to and from text files. The \emph{minstd\_rand0} was also used to ensure correct checkpointing, as other more advanced random number generators in the C++ Standard Library, like \emph{mt19337} (Mersenne Twister) and \emph{minstd\_rand} utilized different serialization implementations so it was not possible to utilize them to send a pre-seeded random number generator as part of a BOINC workunit or return result.

\section{Backpropagation Implementation}
\label{sec:backprop_implementation}

For this work, a fairly standard implementation of stochastic backpropagation was used to train the CNNs, however there were some notable modifications due to the fact that the CNNs were evolved arbitrarily, as opposed to having well defined layers as in most human architected CNNs (\eg, Figure~\ref{fig:example_genomes} provides examples of evolved CNNs). Backpropagation was done as in standard stochastic backpropagation, a forward pass and backward pass with weight updates were performed once per training image, and the order of training images was shuffled before the beginning of each epoch.

\paragraph{Weight Initialization} Apart from the softmax output layer, each node in the CNNs evolved by EXACT used a leaky ReLU activation function with max value 5.5 and a leak of 0.1. Due to this fact, weights are initialized as recmmoneded by He \etal~\cite{he2015delving}, where the variance, $\sigma^2$, of the weights, $w$, input to a neuron is $\sigma^2(w) = \sqrt{\frac{2}{n}}$, where $n$ is the number of weights input to that neuron. After a CNN has been generated by EXACT, when it is initialized a forward pass is made through the graph, and each node calculates the total number of weights input to it, and that value is then used to randomly initialize those weights.

\paragraph{Forward Propagation} As specified in Section~\ref{sec:mutation}, each node in the evolved CNN contains a depth value. By placing all edges into a vector sorted by the depth of their input node, a forward pass through the CNN can be done by propagating forward through each edge in that sorted order as opposed to doing a graph traversal.  Each node keeps track of the number of input edges, and when that many have propagated values forward into the node, it applies the ReLU activation function to each of its neurons, unless it is an output node.

\paragraph{Backward Propagation} Error values can be propagated backwards by each edge in the reverse order of the forward propagation. Weights were updated using Nesterov momentum and L2 Regularization~\cite{bengio2013advances}:

\begin{equation}
v_{prev} = v
\end{equation}
\vspace{-6mm}
\begin{equation}
v = \mu * v - \eta * dw_i
\end{equation}
\vspace{-6mm}
\begin{equation}
w_i +=  -\mu * v_{prev} + (1 + \mu) *v
\end{equation}
\vspace{-6mm}
\begin{equation}
w_i -= w_i * \lambda
\end{equation}

\noindent
where $v_{prev}$ is the previous value for velocity $v$, $\mu$ is the momentum hyperparameter, $\eta$ is the learning rate, $\lambda$ is the L2 regularization weight decay hyperparameter, $dw_i$ is the weight update calculated by error backpropagation for weight $w_i$.





\paragraph{Hyperparameter Updates} At the end of each epoch, the following updates are performed on the three hyperparameters $\mu$, $\eta$, and $\lambda$:

\begin{equation}
\mu = min(\mu * \Delta\mu, \mu_{max})
\end{equation}
\vspace{-6mm}
\begin{equation}
\eta = max(\eta * \Delta\eta, \eta_{min})
\end{equation}
\vspace{-6mm}
\begin{equation}
\lambda = max(\lambda * \Delta\lambda, \lambda_{min})
\end{equation}

\noindent
where $\Delta\eta < 1$, $\Delta\lambda < 1$, and $\Delta\mu > 1$.  These operations decay or increase the hyperparameters to the predefined $\mu_{max}$, $\eta_{min}$ and $\lambda_{min}$ values.

%% file: 04-results.tex
\section{Results}
\label{sec:results}

\subsection{Benchmark Networks}
\label{sec:benchmark_networks}

The implementation of backpropagation, its hyperparameters and possible pre-training can all have significant effects on the resulting performance of a neural network.  As the focus of this work was to demonstrate the effectiveness of EXACT as a evolution strategy for the structure of convolutional neural networks, and due to the fact that the EXACT algorithm can allow for the use of any neural network training strategy, three different neural network architectures of increasing complexity were used as benchmarks to examine how well the evolved networks compare to human designed networks.

\begin{figure}
\begin{center}
\includegraphics[width=0.95\textwidth]{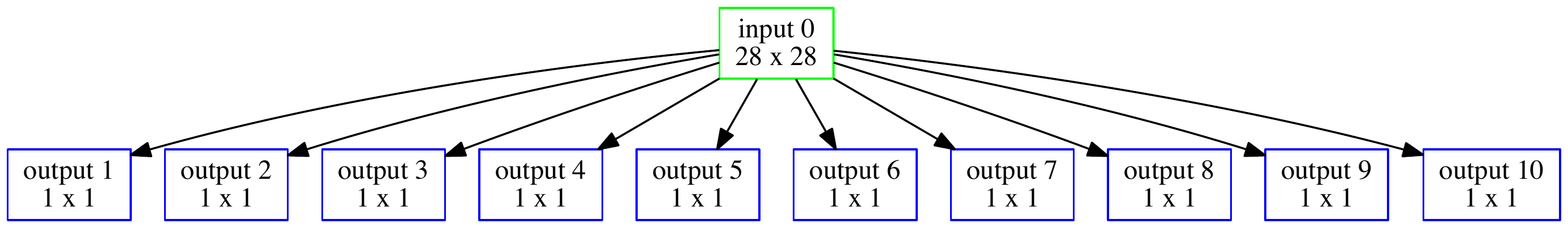}
\caption{\label{fig:one_layer_nn} Benchmark Architecture I: A simple one layer neural network.}
\end{center}
\end{figure}

\begin{figure}
\begin{center}
\includegraphics[width=0.95\textwidth]{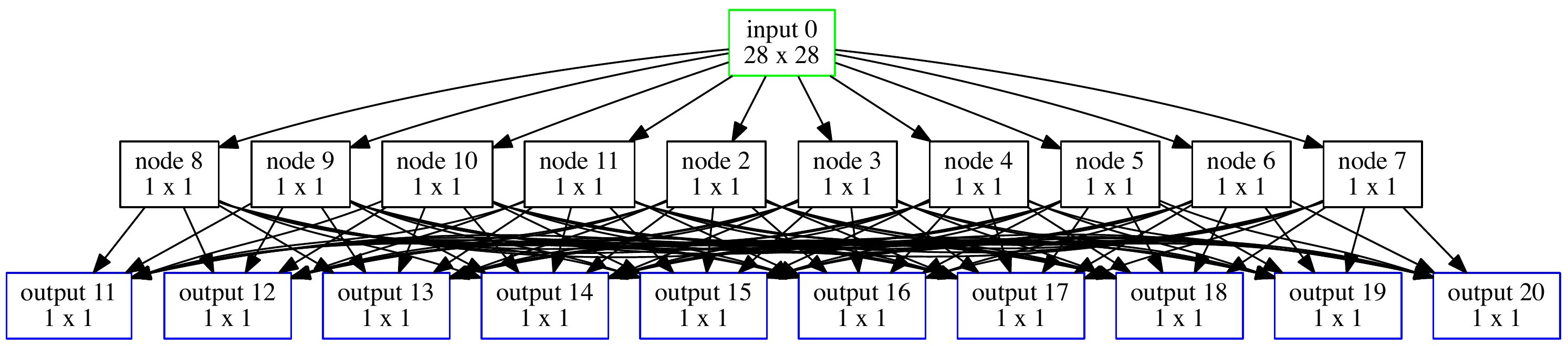}
\caption{\label{fig:two_layer_nn} Benchmark Architecture II: A two layer neural network.}
\end{center}
\end{figure}

\begin{figure}
\begin{center}
\includegraphics[width=0.95\textwidth]{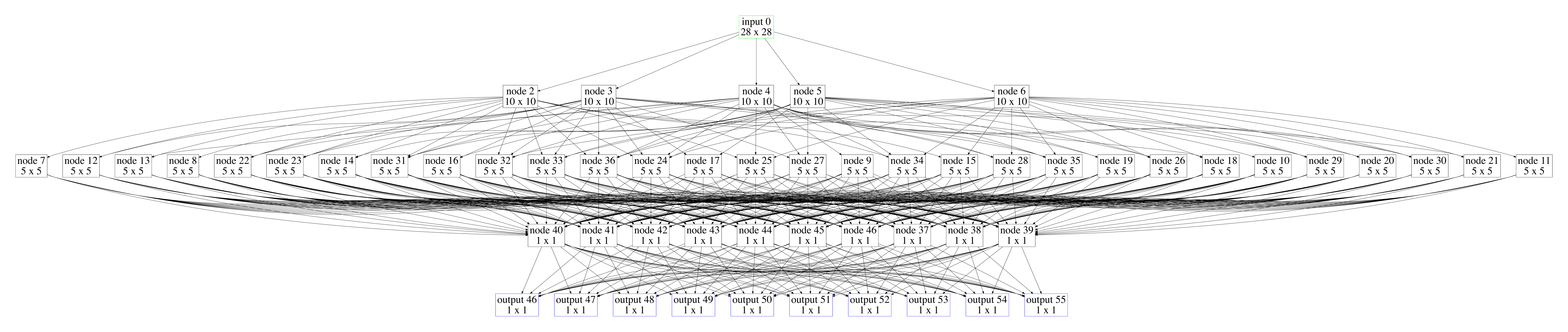}
\caption{\label{fig:modified_lenet} Benchmark Architecture III: A LeNet 5 variant with no max pooling.}
\end{center}
\end{figure}

The first and simplest network was a single layer neural network (see Figure~\ref{fig:one_layer_nn}) where the 28x28 input node feeds 784 weights (one per pixel) into each of the ten output nodes in a softmax layer. The second, a two layer neural network (see Figure~\ref{fig:two_layer_nn}) adds feeds the 28x28 input node feeds 784 weights (one per pixel) into each of 10 nodes in the hidden layer, which are fully connected to the ten output nodes in a final softmax layer.  The third architeecture (see Figure~\ref{fig:modified_lenet}) is based off the popular LeNet 5 architecture~\cite{lecun1998gradient}. The 28x28 input node has five 19x19 filters which feed into five 10x10 nodes in the hidden layer, these feed 80 6x6 filters into 31 5x5 nodes, using all permutations inputs from the 2nd to the 3rd layer.  The third layer is fully connected into 10 nodes, which are fully connected into 10 output nodes in a softmax layer. As the EXACT strategy is not yet evolving max pooling, none of these three networks utilized max pooling layers.

These were trained with a variety of hyperparameters, and finally parameters of momentum $\mu = 0.5$, $\Delta\mu = 1.10$, $\mu_{max} = 0.99$, learning rate $\eta = 0.001$, $\Delta\eta = 0.98$, $eta_{min} = 0.00001$, and weight decay $\lambda = 0.00001$, $\Delta\lambda = 0.98$, and $\lambda_{min} = 0.000001$ were selected. These were chosen in part because they provided the most consistently good results on the three networks, and also provided some of the best results in the most complicated modified LeNet architecture. As the EXACT algorithm trains progressively more complex CNNs, selecting a set of hyperparameters that would be robust to a wide variety of networks that also performed well on complex architectures was of particular importance. 

Table~\ref{table:benchmark_nns} presents the best, average and worst error and predictions on the 60,000 training and 10,000 testing images in the MNIST handwritten digits dataset~\cite{lecun1998mnist}. Results are for 10 different runs initialized with random weights as per Section~\ref{sec:backprop_implementation}. Interestingly, the two layer network was able to find networks with the lowest error, calculated as the sum of the error of each softmax output node for every input image, however the modified LeNet consistently found networks with the best training and test data predictions.  These three networks were used as a baseline to demonstrate the effectiveness of the EXACT algorithm in a manner where the backpropagation algorithm and its hyperparameters were held consistent.

\begin{table*}
\begin{tiny}
\begin{tabular}{|l|r|r|r|r|r|r|r|}
\hline
                    & {\bf Number} & \multicolumn{3}{|c|}{{\bf Training Error}}        & \multicolumn{3}{|c|}{{\bf Testing Error}}\\
\cline{3-8}
{\bf Network}       & {\bf Weights} & {\bf Best}    & {\bf Avg} & {\bf Worst}       & {\bf Best}        & {\bf Avg} & {\bf Worst} \\
\hline
\hline
One Layer           & 7840      &   16222.59      & 16637.26      & 17457.19          & 2643.16           & 2686.82       & 2792.68 \\
\hline
Two Layer           & 8260      & 7041.03       & 8063.04       & 9084.34           & 1186.11           & 1331.50       & 1499.41 \\
\hline
Modified LeNet      & 12285     & 7994.90       & 8556.57       & 9484.93           & 1325.92           & 1408.41       & 1585.14 \\
\hline
\hline
                    & {\bf Number} & \multicolumn{3}{|c|}{{\bf Training Predictions}} & \multicolumn{3}{|c|}{{\bf Testing Predictions}} \\
\cline{3-8}
{\bf Network}       & {\bf Weights} & {\bf Best}    & {\bf Avg} & {\bf Worst}       & {\bf Best}        & {\bf Avg} & {\bf Worst} \\
\hline
\hline
One Layer           & 7840      & 92.02\%       & 91.04\%       & 80.67\%   & 91.97\%   & 91.10\%   &   90.09\%  \\
\hline
Two Layer           & 8260      & 96.66\%           & 96.66\%       & 94.77\%           & 96.66\%       & 95.90\%       & 94.63\%      \\
\hline
Modified LeNet      & 12285     & 97.27\%           & 96.83\%       & 96.09\%           & 97.19\%       & 96.79\%       & 95.90\%      \\
\hline
\end{tabular}
\caption{\label{table:benchmark_nns} Benchmark Neural Network Error and Prediction Rates}
\end{tiny}
\end{table*}

\begin{table*}
\begin{tiny}
\begin{tabular}{|l|r|r|r|r|r|r|r|}
\hline
                    & {\bf Avg. Num.} & \multicolumn{3}{|c|}{{\bf Training Error}}        & \multicolumn{3}{|c|}{{\bf Testing Error}} \\
\cline{3-8}
{\bf Network}       & {\bf Weights} & {\bf Best}    & {\bf Avg} & {\bf Worst}       & {\bf Best}        & {\bf Avg} & {\bf Worst} \\
\hline
\hline
Randomized          & 25,603.35 & 3,494.54 & 3,742.22 & 3,825.23 & 544.26 & 603.17 & 682.75 \\
\hline
Epigenetic          & 23862.65 & 3,644.30 & 3,909.88 & 3,991.73 & 594.13 & 657.48 & 710.25 \\
\hline
\hline
                    & {\bf Avg. Num.} & \multicolumn{3}{|c|}{{\bf Training Predictions}} & \multicolumn{3}{|c|}{{\bf Testing Predictions}} \\
\cline{3-8}
{\bf Network}       & {\bf Weights} & {\bf Best}    & {\bf Avg} & {\bf Worst}       & {\bf Best}        & {\bf Avg} & {\bf Worst} \\
\hline
\hline
Randomized          & 25,603.35 & 97.75\% & 97.33\% & 97.05\% & 97.89\% & 97.40\% & 96.98\% \\
\hline
Epigenetic          & 23862.65 & 98.42\% & 97.98\% & 97.48\% & 98.32\% & 97.87\% & 97.28\% \\
\hline
\end{tabular}
\caption{\label{table:exact_nns} Error and Prediction Rates for the top 20 CNNs in each EXACT search.}
\end{tiny}
\end{table*}

\subsection{Evolved Neural Networks}
\label{sec:evolved_neural_networks}

\paragraph{EXACT Hyperparameters} Two simultaneous EXACT searches were performed, one using epigenetic weight initialization and the other using randomized weight initialization. They both had a population size of 100, a crossover rate of 20\% (which entails a mutation rate of 80\%) and 3 mutation operations were performed for each genome mutated. Edges were disabled $\frac{1}{11}$th of the time, enabled $\frac{2}{11}$ths of the time, split $\frac{2}{11}$ths of the time, and added $\frac{4}{11}$ths of the time; filter sizes were changed in both x and y dimensions $\frac{1}{11}$th of the time, and in just the x dimension or y dimension each $\frac{1}{22}$nd of the time. These parameters were chosen due to experience testing the EXACT algorithm on smaller subests of the training data using the multithreaded and MPI implementations. The CNNs were trained with the same hyperparameters as the benchmark CNNs.

\begin{figure*}
\includegraphics[width=0.32\textwidth]{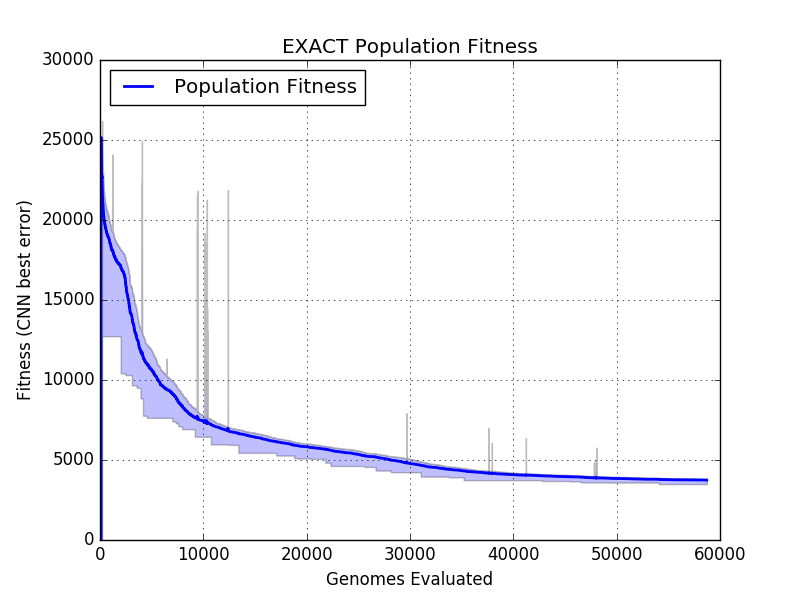}
\includegraphics[width=0.32\textwidth]{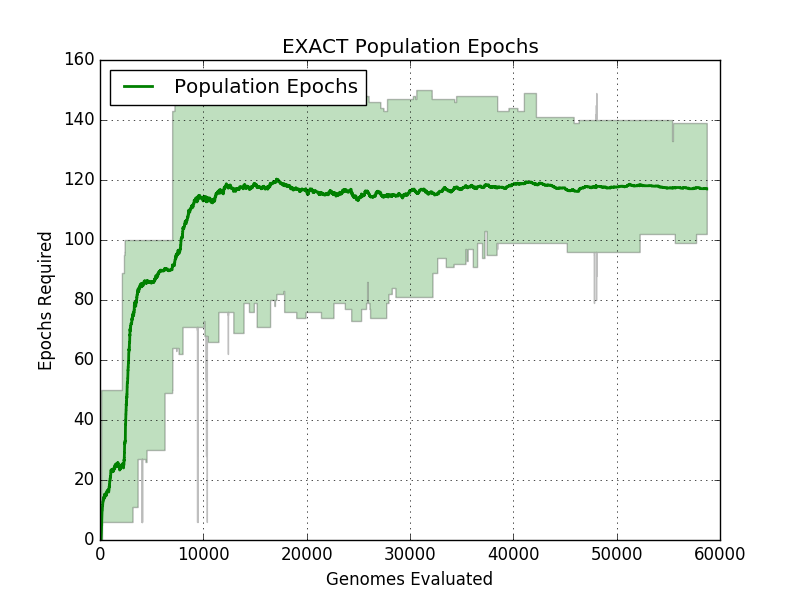}
\includegraphics[width=0.32\textwidth]{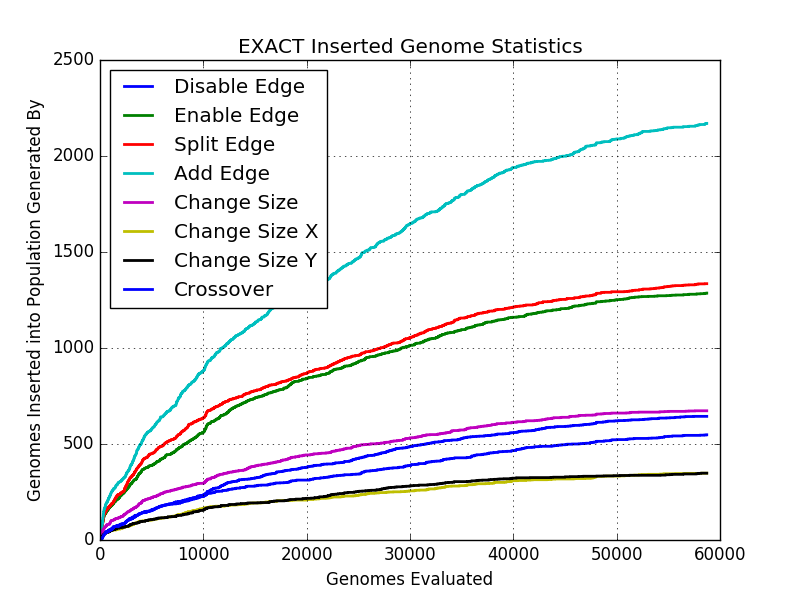}
\caption{\label{fig:reset_progress} Progress of the search which used randomized weight initialization, showing the average and range of fitness in the population (left), the average and range of which epoch the CNN found minimal training error (center), and how often a genome was inserted into the population having been generated by crossover or a particular mutation operation (right).}
\end{figure*}

\begin{figure*}
\includegraphics[width=0.32\textwidth]{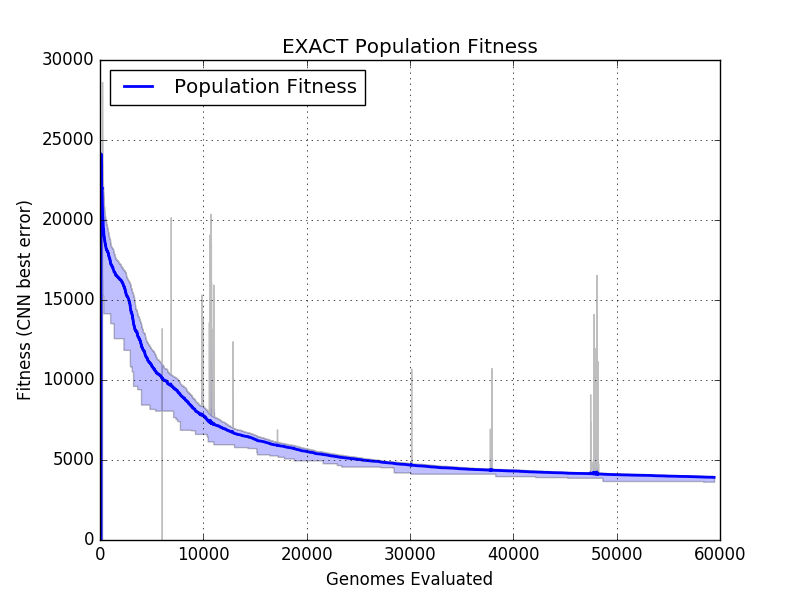}
\includegraphics[width=0.32\textwidth]{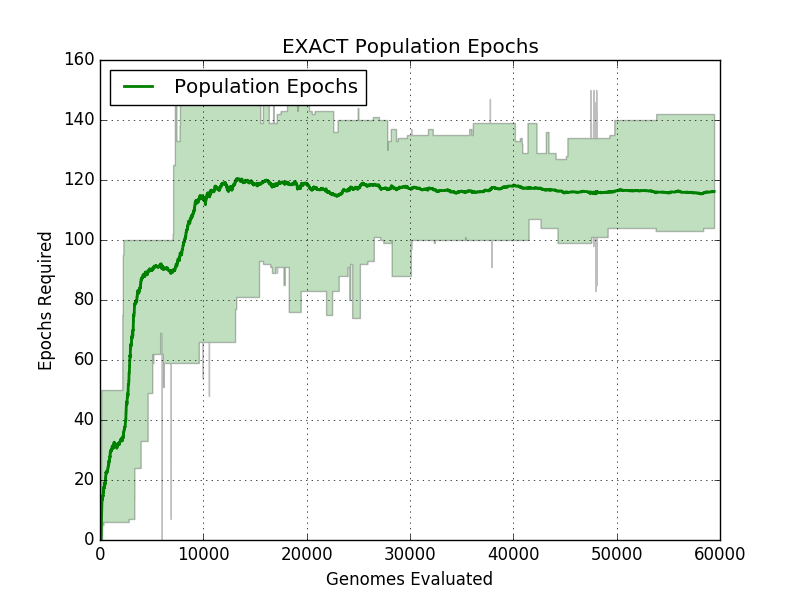}
\includegraphics[width=0.32\textwidth]{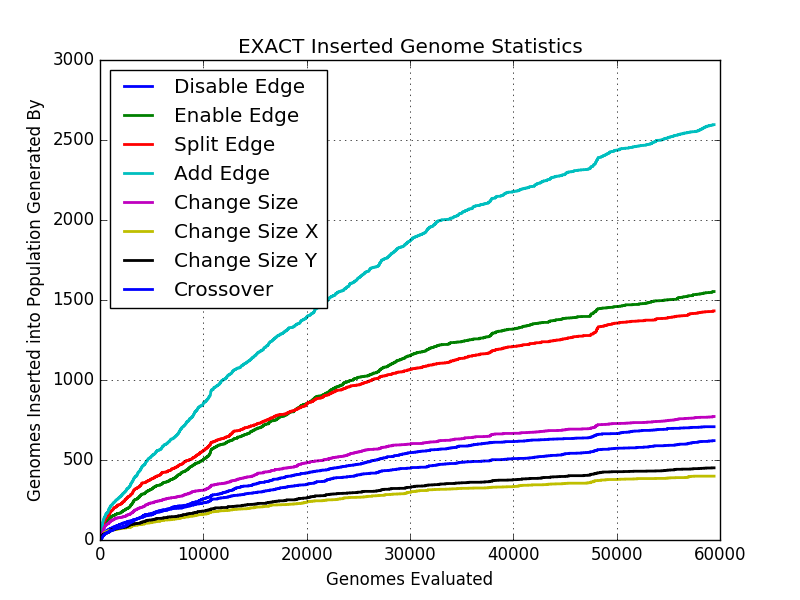}
\caption{\label{fig:noreset_progress} Progress of the search which used epigenetic weight initialization, showing the average and range of fitness in the population (left), the average and range of which epoch the CNN found minimal training error (center), and how often a genome was inserted into the population having been generated by crossover or a particular mutation operation (right).}
\end{figure*}

\paragraph{Search Progress} Figures~\ref{fig:reset_progress} and~\ref{fig:noreset_progress} show the progress of the two EXACT searches in terms of the range of fitness of the populations, the range of how many epochs it took each genome to find its minimum training error, and how many genomes were inserted into the population having been generated with crossover and the different mutation operations.  Spikes occured when some of the BOINC daemons handling the EXACT searches had to be restarted. The CNNs were initially allowed to train for 50 epochs, which was increased to 100 epochs and then finally 150 epochs as it became apparent that the CNNs could reach lower training error with more epochs.

Both searches progressed at nearly the same rate in terms of finding CNNs with low training error and also required a similar amount of epochs to reach a minimum training error. This was in contrast to the intuition that the inheriting good weights from the parents would provide a good starting weights for the CNNs, potentially reducing training time and allowing them to find better weights.  However, as shown in Table~\ref{table:exact_nns}, the CNNs evolved using epigentic weight initialization on average found solutions with noticeably better training and testing predictions, even as randomized weight initialization found CNNs with on average better training and testing error.  From these results it does appear that epigenetic weight initialization is having some positive benefit on CNN training, and begs further investigation.

\begin{figure*}
\begin{subfigure}[b]{0.99\textwidth}
\begin{center}
\includegraphics[width=0.99\textwidth]{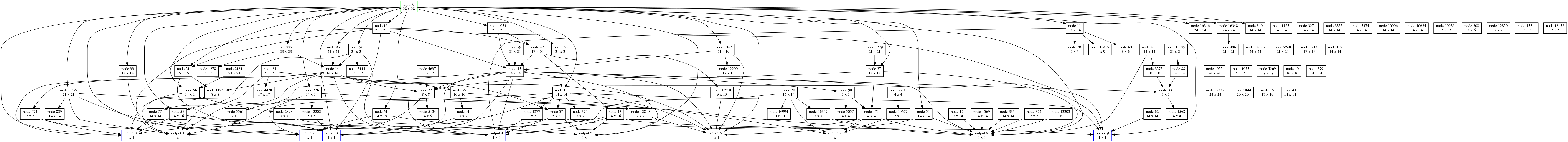}
\caption{\label{fig:reset_genome_57302} Genome 57302 had the lowest training error for the randomized weight search. This network had a training error of 3,494.54, test error of 603.30, training accuracy of 97.75\% and test accuracy of 97.58\%.}
\end{center}
\end{subfigure}

\vspace{4mm}
\begin{subfigure}[b]{0.99\textwidth}
\begin{center}
\includegraphics[width=0.99\textwidth]{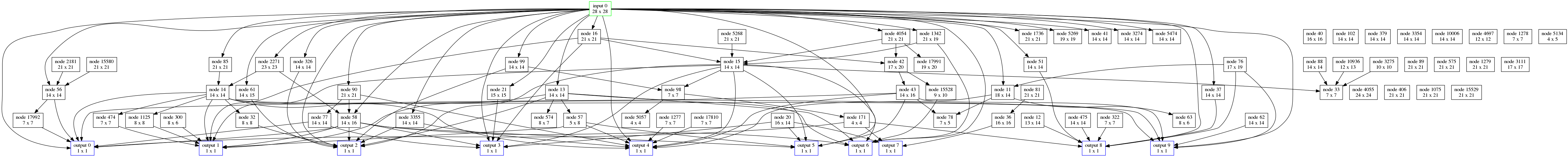}
\caption{\label{fig:reset_genome_46823} Genome 46823 had the best training accuracy for the randomized weight search. This network had a training error of 3,686.73, test error of 583.89, training accuracy of 97.66\% and test accuracy of 97.89\%.}
\end{center}
\end{subfigure}

\vspace{4mm}
\begin{subfigure}[b]{0.99\textwidth}
\begin{center}
\includegraphics[width=0.99\textwidth]{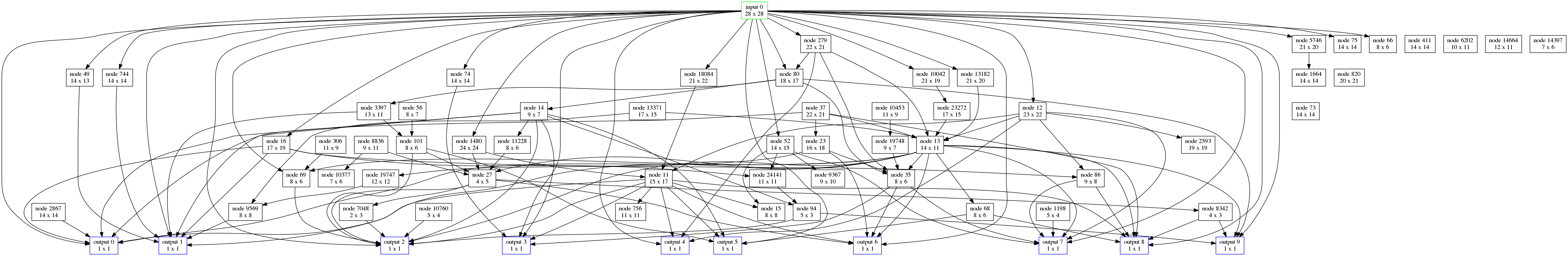}
\caption{\label{fig:noreset_genome_59920} Genome 59920 had the lowest training error for the epigenetic weight search. This network had a training error of 3,644.30, test error of 594.13, training accuracy of 97.81\% and test accuracy of 97.92\%.}
\end{center}
\end{subfigure}

\vspace{4mm}
\begin{subfigure}[b]{0.99\textwidth}
\begin{center}
\includegraphics[width=0.99\textwidth]{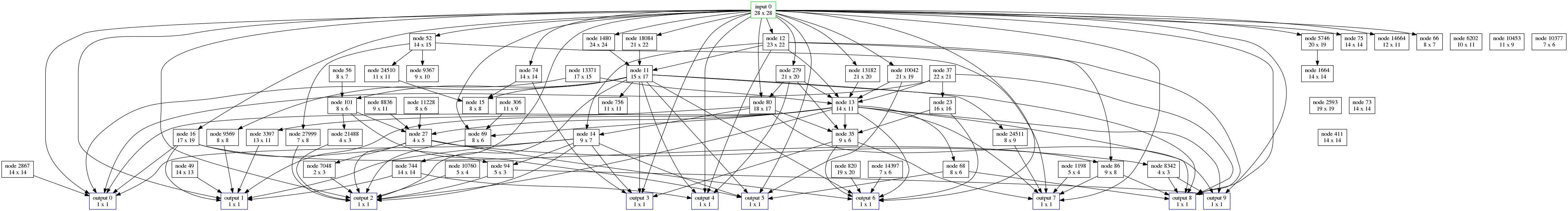}
\caption{\label{fig:noreset_genome_59455} Genome 59455 had the best training accuracy for the epigenetic weight search. This network had a training error of 3,830.97, test error of 633.95, training accuracy of 98.42\% and test accuracy of 98.32\%.}
\end{center}
\end{subfigure}

\caption{\label{fig:example_genomes} These four CNN genomes represent some of the best found CNNs evolved by the EXACT searches. Disabled edges are not shown, however vestigial filters and edges are shown. Checkpoints for these genomes can be found in the EXACT github repository to be run over the MNIST dataset for validation.}
\end{figure*}

\paragraph{Evolved Genomes} Figure~\ref{fig:example_genomes} shows some of the best CNNs evolved by the two searches. These networks are quite interesting in that they are highly different from the highly structured CNNs found seen in literature~\cite{lecun1998gradient,krizhevsky2012imagenet,simonyan2014very,szegedy2015going,he2016deep}. Even so, compared to the hand designed benchmark networks, they still were able to find CNNs that trained to significantly lower training and test error, while making strong improvements in training and testing accuracy.  The networks also show vestigial filters and edges, resulting from the crossover and edge disable mutation operator. Genomes 57302 and 46823 (Figures~\ref{fig:reset_genome_57302} and Figure~\ref{fig:reset_genome_46823}) are particularly interesting in that they shows a large number of disconnected or otherwise unused nodes. It is also interesting that there appear to be some filters which have larger numbers of input and output edges, which begin to bear some resemblence to neurons seen in the cerebral cortex.

\paragraph{Hyperparameter Investigation} Figures~\ref{fig:reset_progress} and~\ref{fig:noreset_progress} show how many genomes were inserted into the population over time that were produced by crossover and the various mutation operators. While these mostly followed the probability of a genome being generated by those operators, they did appear to change over time. In particular, for the search with epigenetic weight initialization, initially splitting edges more frequently resulted in a genome being inserted into the population than enabling an edge, however after time enabling edges became more successful. Likewise, for both searches there became a point where crossover became to more beneficial as it became more successful than disabling edges.  These results suggest that adapting the EXACT hyperparamters (as has been done in recent work~\cite{mikkulainen2017codeepneat,real2017evolution} may be able to provide further improvements to the results.

%% file: 05-discussion.tex
\section{Discussion and Future Work}
\label{sec:discussion}

A novel algorithm for the evolution of arbitrarily structured CNNs called evolutionary exploration of augmenting convolutional topoligies (EXACT) has been presented, which to the author's knowlege is the first capable of performing this task.  In order to overcome the computational demands of evolving large numbers of CNNs, it was implemented as part of the Citizen Science Grid, a BOINC volunteer computing project. Over 4,500 volunteered compute hosts were used to train over 120,000 evolved CNNs during a period of 2 months, which resulted in CNNs reaching 98.32\% test accuracy on the MNIST handwritten digits data set, and showing significantly improved training and testing error over human designed benchmark nerual networks. The evolved neural networks show significant differences from the highly structued human designed CNNs found in the literature, containing interesting structures which bear some similarities to biological neurons. The author hopes that these may provide some new insights to the machine learning community in the development of new CNN architectures.

This work opens the door for significant future work. In particular, the EXACT algorithm does not yet evolve pooling layers. This can be done by having each filter perform a pooling operation of an arbrarity size, which can be mutated by additional mutation operations. The algorithm also currently only supports 2 dimensional input and filters, and will need to be updated to utilize 3 dimensional inputs and filters so that it can evolve CNNs for color data sets such as the CIFAR and TinyImage datasets~\cite{krizhevsky2009learning,torralba200880}.

The use of epigenetic weight initialization, where child CNNs reuse trained weights from their parents has shown potential for improving the CNNs evolved by EXACT, however it did not seem to reduce the number of epochs required by backpropagation to find a minimal training error.  This may be because different CNN training hyperparameters may provide more effective for these CNNs.  It may also be possible to evolve the hyperparameters used to train the CNNs along with their structure for improved results as done by other recent work~\cite{mikkulainen2017codeepneat,real2017evolution}. Pretraining the CNNs using restricted boltzmann machines~\cite{hinton2006fast} has the potential to even further improve accuracy of the trained CNNs, and may be potentially combined with epigenetic pretraining. Lastly, EXACT evolves and trains a large number of CNNs in each search, which provides an opportunity to determine how robust various CNN training techniques are and to see if these methods have any effect on the structure of the CNNs involved.

%% file: 06-acknowledgements.tex
\section*{Acknowledgments}

We would like to give a special thanks to all the volunteers on the Citizen Science Grid who generously provided their CPU cycles to this research. This work has been partially supported by the National Science Foundation under Grant Number 1319700. Any opinions, findings, and conclusions or recommendations expressed in this material are those of the authors and do not necessarily reflect the views of the National Science Foundation.

%% file: main.bbl
\begin{thebibliography}{10}

\bibitem{anderson-volunteer-computing-2010}
D.~P. Anderson.
\newblock Volunteer computing: the ultimate cloud.
\newblock {\em Crossroads}, 16(3):7--10, 2010.

\bibitem{anderson_boinc_2005}
D.~P. Anderson, E.~Korpela, and R.~Walton.
\newblock High-performance task distribution for volunteer computing.
\newblock In {\em e-Science}, pages 196--203. IEEE Computer Society, 2005.

\bibitem{bastien2012theano}
F.~Bastien, P.~Lamblin, R.~Pascanu, J.~Bergstra, I.~Goodfellow, A.~Bergeron,
  N.~Bouchard, D.~Warde-Farley, and Y.~Bengio.
\newblock Theano: new features and speed improvements.
\newblock {\em arXiv preprint arXiv:1211.5590}, 2012.

\bibitem{bengio2013advances}
Y.~Bengio, N.~Boulanger-Lewandowski, and R.~Pascanu.
\newblock Advances in optimizing recurrent networks.
\newblock In {\em Acoustics, Speech and Signal Processing (ICASSP), 2013 IEEE
  International Conference on}, pages 8624--8628. IEEE, 2013.

\bibitem{bergstra2011theano}
J.~Bergstra, F.~Bastien, O.~Breuleux, P.~Lamblin, R.~Pascanu, O.~Delalleau,
  G.~Desjardins, D.~Warde-Farley, I.~Goodfellow, A.~Bergeron, et~al.
\newblock Theano: Deep learning on gpus with python.
\newblock In {\em NIPS 2011, BigLearning Workshop, Granada, Spain}, volume~3.
  Citeseer, 2011.

\bibitem{boincstats-2017}
BoincStats.
\newblock {BOINC} stats, 2017.
\newblock http://boincstats.com/.

\bibitem{desell-analysis-massive-eas-2010}
T.~Desell, D.~Anderson, M.~Magdon-Ismail, B.~S. Heidi~Newberg, and C.~Varela.
\newblock An analysis of massively distributed evolutionary algorithms.
\newblock In {\em The 2010 IEEE congress on evolutionary computation (IEEE CEC
  2010)}, Barcelona, Spain, July 2010.

\bibitem{desell-evostar-2015}
T.~Desell, S.~Clachar, J.~Higgins, and B.~Wild.
\newblock Evolving deep recurrent neural networks using ant colony
  optimization.
\newblock In G.~Ochoa and F.~Chicano, editors, {\em Evolutionary Computation in
  Combinatorial Optimization}, volume 9026 of {\em Lecture Notes in Computer
  Science}, pages 86--98. Springer International Publishing, 2015.

\bibitem{gomez2008accelerated}
F.~Gomez, J.~Schmidhuber, and R.~Miikkulainen.
\newblock Accelerated neural evolution through cooperatively coevolved
  synapses.
\newblock {\em Journal of Machine Learning Research}, 9(May):937--965, 2008.

\bibitem{gridcoin-2017}
Gridcoin.
\newblock What is gridcoin?, 2017.
\newblock http://www.gridcoin.us/.

\bibitem{he2015delving}
K.~He, X.~Zhang, S.~Ren, and J.~Sun.
\newblock Delving deep into rectifiers: Surpassing human-level performance on
  imagenet classification.
\newblock In {\em Proceedings of the IEEE international conference on computer
  vision}, pages 1026--1034, 2015.

\bibitem{he2016deep}
K.~He, X.~Zhang, S.~Ren, and J.~Sun.
\newblock Deep residual learning for image recognition.
\newblock In {\em Proceedings of the IEEE Conference on Computer Vision and
  Pattern Recognition}, pages 770--778, 2016.

\bibitem{heien2009pymw}
E.~M. Heien, Y.~Takata, K.~Hagihara, and A.~Kornafeld.
\newblock Pymw-a python module for desktop grid and volunteer computing.
\newblock In {\em Parallel \& Distributed Processing, 2009. IPDPS 2009. IEEE
  International Symposium on}, pages 1--7. IEEE, 2009.

\bibitem{hinton2012deep}
G.~Hinton, L.~Deng, D.~Yu, G.~E. Dahl, A.-r. Mohamed, N.~Jaitly, A.~Senior,
  V.~Vanhoucke, P.~Nguyen, T.~N. Sainath, et~al.
\newblock Deep neural networks for acoustic modeling in speech recognition: The
  shared views of four research groups.
\newblock {\em IEEE Signal Processing Magazine}, 29(6):82--97, 2012.

\bibitem{hinton2006fast}
G.~Hinton, S.~Osindero, and Y.-W. Teh.
\newblock A fast learning algorithm for deep belief nets.
\newblock {\em Neural computation}, 18(7):1527--1554, 2006.

\bibitem{jia2014caffe}
Y.~Jia, E.~Shelhamer, J.~Donahue, S.~Karayev, J.~Long, R.~Girshick,
  S.~Guadarrama, and T.~Darrell.
\newblock Caffe: Convolutional architecture for fast feature embedding.
\newblock In {\em Proceedings of the ACM International Conference on
  Multimedia}, pages 675--678. ACM, 2014.

\bibitem{karpathy2014large}
A.~Karpathy, G.~Toderici, S.~Shetty, T.~Leung, R.~Sukthankar, and L.~Fei-Fei.
\newblock Large-scale video classification with convolutional neural networks.
\newblock In {\em Proceedings of the IEEE conference on Computer Vision and
  Pattern Recognition}, pages 1725--1732, 2014.

\bibitem{kim2014convolutional}
Y.~Kim.
\newblock Convolutional neural networks for sentence classification.
\newblock {\em arXiv preprint arXiv:1408.5882}, 2014.

\bibitem{koutnik2014evolving}
J.~Koutn{\'\i}k, J.~Schmidhuber, and F.~Gomez.
\newblock Evolving deep unsupervised convolutional networks for vision-based
  reinforcement learning.
\newblock In {\em Proceedings of the 2014 Annual Conference on Genetic and
  Evolutionary Computation}, pages 541--548. ACM, 2014.

\bibitem{krizhevsky2009learning}
A.~Krizhevsky and G.~Hinton.
\newblock Learning multiple layers of features from tiny images.
\newblock {\em Computer Science Department, University of Toronto, Tech. Rep},
  2009.

\bibitem{krizhevsky2012imagenet}
A.~Krizhevsky, I.~Sutskever, and G.~E. Hinton.
\newblock Imagenet classification with deep convolutional neural networks.
\newblock In F.~Pereira, C.~Burges, L.~Bottou, and K.~Weinberger, editors, {\em
  Advances in Neural Information Processing Systems 25}, pages 1097--1105.
  Curran Associates, Inc., 2012.

\bibitem{lecun1998gradient}
Y.~LeCun, L.~Bottou, Y.~Bengio, and P.~Haffner.
\newblock Gradient-based learning applied to document recognition.
\newblock {\em Proceedings of the IEEE}, 86(11):2278--2324, 1998.

\bibitem{lecun1998mnist}
Y.~LeCun, C.~Cortes, and C.~J. Burges.
\newblock The mnist database of handwritten digits, 1998.

\bibitem{mikkulainen2017codeepneat}
R.~Miikkulainen, J.~Liang, E.~Meyerson, A.~Rawal, D.~Fink, O.~Francon, B.~Raju,
  H.~Shahrzad, A.~Navruzyan, N.~Duffy, and B.~Hodjat.
\newblock Evolving deep neural networks.
\newblock {\em arXiv preprint arXiv:1703.00548}, 2017.

\bibitem{rentala-vmwrapper}
J.~Rantala.
\newblock {VMWrapper}, 2017.
\newblock http://boinc.berkeley.edu/trac/wiki/VmApps.

\bibitem{real2017evolution}
E.~Real, S.~Moore, A.~Selle, S.~Saxena, Y.~L. Suematsu, Q.~Le, and A.~Kurakin.
\newblock Large-scale evolution of image classifiers.
\newblock {\em arXiv preprint arXiv:1703.01041}, 2017.

\bibitem{salama2014novel}
K.~Salama and A.~M. Abdelbar.
\newblock A novel ant colony algorithm for building neural network topologies.
\newblock In {\em Swarm Intelligence}, pages 1--12. Springer, 2014.

\bibitem{simonyan2014very}
K.~Simonyan and A.~Zisserman.
\newblock Very deep convolutional networks for large-scale image recognition.
\newblock {\em arXiv preprint arXiv:1409.1556}, 2014.

\bibitem{stanley2002evolving}
K.~Stanley and R.~Miikkulainen.
\newblock Evolving neural networks through augmenting topologies.
\newblock {\em Evolutionary computation}, 10(2):99--127, 2002.

\bibitem{stanley2009hypercube}
K.~O. Stanley, D.~B. D'Ambrosio, and J.~Gauci.
\newblock A hypercube-based encoding for evolving large-scale neural networks.
\newblock {\em Artificial life}, 15(2):185--212, 2009.

\bibitem{szegedy2015going}
C.~Szegedy, W.~Liu, Y.~Jia, P.~Sermanet, S.~Reed, D.~Anguelov, D.~Erhan,
  V.~Vanhoucke, and A.~Rabinovich.
\newblock Going deeper with convolutions.
\newblock In {\em Proceedings of the IEEE Conference on Computer Vision and
  Pattern Recognition}, pages 1--9, 2015.

\bibitem{torralba200880}
A.~Torralba, R.~Fergus, and W.~T. Freeman.
\newblock 80 million tiny images: A large data set for nonparametric object and
  scene recognition.
\newblock {\em Pattern Analysis and Machine Intelligence, IEEE Transactions
  on}, 30(11):1958--1970, 2008.

\bibitem{xie2017geneticcnn}
L.~Xie and A.~Yuille.
\newblock Genetic cnn.
\newblock {\em arXiv preprint arXiv:1703.01513}, 2017.

\bibitem{zoph2016neural}
B.~Zoph and Q.~V. Le.
\newblock Neural architecture search with reinforcement learning.
\newblock {\em arXiv preprint arXiv:1611.01578}, 2016.

\end{thebibliography}
